# O-CNN: Octree-based Convolutional Neural Networks for 3D Shape Analysis

PENG-SHUAI WANG, Tsinghua University and Microsoft Research Asia
YANG LIU, Microsoft Research Asia
YU-XIAO GUO, University of Electronic Science and Technology of China and Microsoft Research Asia
CHUN-YU SUN, Tsinghua University and Microsoft Research Asia
XIN TONG, Microsoft Research Asia

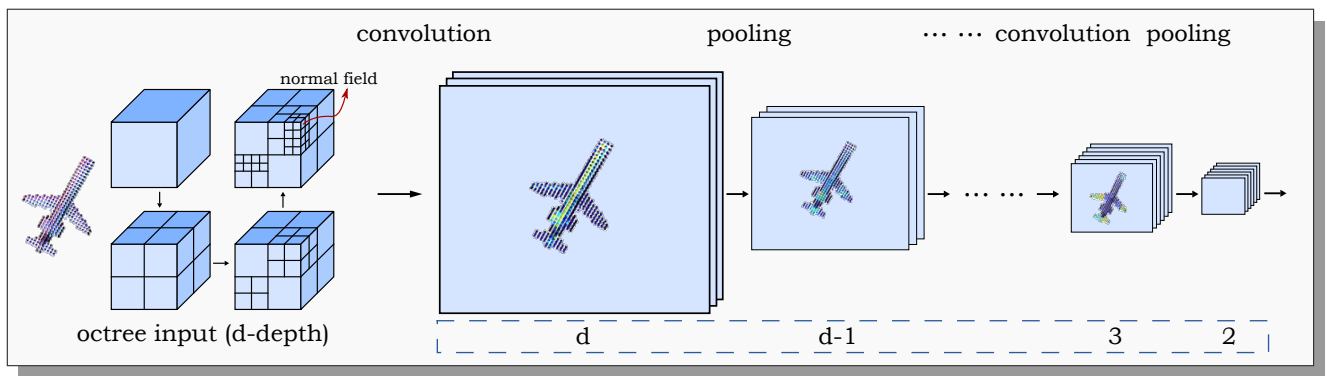

Fig. 1. An illustration of our octree-based convolutional neural network (O-CNN). Our method represents the input shape with an octree and feeds the averaged normal vectors stored in the finest leaf octants to the CNN as input. All the CNN operations are efficiently executed on the GPU and the resulting features are stored in the octree structure. Numbers inside the blue dashed square denote the depth of the octants involved in computation.

We present *O-CNN*, an Octree-based Convolutional Neural Network (CNN) for 3D shape analysis. Built upon the octree representation of 3D shapes, our method takes the average normal vectors of a 3D model sampled in the finest leaf octants as input and performs 3D CNN operations on the octants occupied by the 3D shape surface. We design a novel octree data structure to efficiently store the octant information and CNN features into the graphics memory and execute the entire O-CNN training and evaluation on the GPU. O-CNN supports various CNN structures and works for 3D shapes in different representations. By restraining the computations on the octants occupied by 3D surfaces, the memory and computational costs of the O-CNN grow quadratically as the depth of the octree increases, which makes the 3D CNN feasible for high-resolution 3D models. We compare the performance of the O-CNN with other existing 3D CNN solutions and demonstrate the efficiency and efficacy of O-CNN in three shape analysis tasks, including object classification, shape retrieval, and shape segmentation.

## 1 INTRODUCTION

With recent advances in low-cost 3D acquisition devices and 3D modeling tools, the amount of 3D models created by end users has been increasing quickly. Analyzing and understanding these 3D shapes, as for classification, segmentation, and retrieval, have become more and more important for many graphics and vision applications. A key technique for these shape analysis tasks is to extract features of 3D models that can sufficiently characterize their shapes and parts.

In the computer vision field, convolutional neural networks (CNNs) are widely used for image feature extraction and have demonstrated their advantages over manually-crafted solutions in most image analysis and understanding tasks. However, it is a non-trivial task to adapt a CNN designed for regularly sampled 2D images to 3D shapes modeled by irregular triangle meshes or point clouds. A set of methods convert the 3D shapes to regularly sampled representations and apply a CNN to them. Voxel-based methods [Maturana and Scherer 2015; Wu et al. 2015] rasterize a 3D shape as an indicator function or distance function sampled over dense voxels and apply a 3D CNN over the entire 3D volume. Since the memory and computation cost grow cubically as the voxel resolution increases, these methods become prohibitively expensive for high-resolution voxels. Manifold-based methods [Boscaini et al. 2015, 2016; Masci et al. 2015; Sinha et al. 2016] perform CNN computations over the features defined on a 3D mesh manifold. These methods require smooth manifold surfaces as input and are sensitive to noise and large distortion, which makes them unsuitable for the non-manifold 3D models in many 3D shape repositories. Multiple-view based approaches [Bai et al. 2016; Shi et al. 2015; Su et al. 2015] render the 3D shape into a set of 2D images observed from different views and feed the stacked images to the CNN. However, it is unclear how to determine the view positions to cover full 3D shapes and avoid self-occlusions.

We present an octree-based convolutional neural network, named *O-CNN*, for 3D shape analysis. The key idea of our method is to represent the 3D shapes with octrees and perform 3D CNN operations only on the sparse octants occupied by the boundary surfaces of 3D shapes. To this end, the O-CNN takes the average normal vectors of a 3D model sampled in the finest leaf octants as input and computes features for the finest level octants. After pooling, the features are down-sampled to the parent octants in the next coarser level and are fed into the next O-CNN layer. This process is repeated until all O-CNN layers are evaluated.

The main technical challenge of the O-CNN is to parallelize the O-CNN computations defined on the sparse octants so that they can be efficiently executed on the GPU. To this end, we design a novel octree structure that stores the features and associated octant information into the graphics memory for supporting all CNN operations on the GPU. In particular, we pack the features and data of sparse octants at each depth as continuous arrays. A label buffer is introduced to find the correspondence between the features at different levels for efficient convolution and pooling operations. To efficiently compute 3D convolutions with an arbitrary kernel size, we build a hash table to quickly construct the local neighborhood volume of eight sibling octants and compute the 3D convolutions of these octants in parallel. With the help of this octree structure, the entire training and evaluation process can be efficiently executed on the GPU.

The O-CNN provides a generic and efficient CNN solution for 3D shape analysis. It supports various CNN structures and works for 3D shapes in different representations. By restraining the CNN computations and features on sparse octants of the 3D shape boundaries, the memory and computation costs of O-CNN grow quadratically as the octree depth increases, which makes it efficient for analyzing high-resolution 3D models. To demonstrate the efficiency of the O-CNN, we construct an O-CNN with basic CNN layers as shown in Figure 1. We train this O-CNN model with 3D shape datasets and refine the O-CNN models with different back-ends for three shape analysis tasks, including object classification, shape retrieval, and shape segmentation. Compared to existing 3D CNN solutions, our method achieves comparable or better accuracy with much less computational and memory costs in all three shape analysis tasks. We also evaluate the performance of the O-CNN with different octree depths in object classification and demonstrate the efficiency and efficacy of the O-CNN for analyzing high-resolution 3D shapes.

## 2 RELATED WORK

In this section, we first review existing CNN approaches and other deep learning methods for 3D shape analysis. Then we discuss the GPU based octree structures used in different graphics applications.

*CNNs for 3D shape analysis.* A set of CNN methods have been presented for 3D shape analysis. We classify these approaches into several classes according to the 3D shape representation used in each solution.

*Voxel-based methods.* model the 3D shape as a function sampled on voxels and define a 3D CNN over voxels for shape analysis. Wu *et al.* [2015] proposed 3D ShapeNets for object recognition and shape completion. Maturana and Scherer [2015] improve 3D ShapeNets with fewer input parameters defined in each voxel. These full-voxel-based methods are limited to low resolutions like $30^3$ due to the high memory and computational cost.

To reduce the computational cost of full-voxel based methods, Graham [2015] proposes the 3D sparse CNNs that apply CNN operations to active voxels and activate only the neighboring voxels inside the convolution kernel. However, the method quickly becomes less efficient as the number of convolution layers between the pooling layers increases. For a CNN with deep layers and a large kernel size, the computational and memory cost of this method is still high. Riegler *et al.* [2017] combine the octree and a grid structure to support high-resolution 3D CNNs. Their method limits the 3D CNN to the interior volume of 3D shapes and becomes less efficient than the full-voxel-based solution when the volume resolution is lower than $64^3$. Our method limits the 3D CNN to the octants of the 3D shape boundaries and leverages a novel octree structure for efficiently training and evaluating the O-CNN on the GPU.

*Manifold-based methods.* perform CNN operations over the geometric features defined on a 3D mesh manifold. Some methods parameterize the 3D surfaces to 2D patches [Boscaini et al. 2015, 2016; Masci et al. 2015; Sinha et al. 2016] or geometry images and feed the regularly sampled feature images into a 2D CNN for shape analysis. Other methods extend the CNN to the graphs defined by irregular triangle meshes [Bronstein et al. 2017]. Although these methods are robust to isometric deformation of 3D shapes, they are constrained to smooth manifold meshes. The local features used in these methods are always computationally expensive. A good survey of these techniques can be found in [Bronstein et al. 2017].

*Multiview-based methods.* represent the 3D shape with a set of images rendered from different views and take the image stacks as the input of a 2D CNN for shape analysis [Bai et al. 2016; Qi et al. 2016; Su et al. 2015]. Although these methods can directly exploit the image-based CNNs for 3D shape analysis and handle high-resolution inputs, it is unclear how to determine the number of views and distribute the views to cover the 3D shape while avoiding self-occlusions. Our method is based on the octree representation and avoids the view selection issue. It can also handle high-resolution inputs and achieve similar performance and accuracy to multiview-based methods.

*Deep learning for 3D shape analysis.* Besides the CNN, other deep learning methods have also been proposed for 3D shape analysis. For shape segmentation, Guo *et al.* [2015] extract low-level feature vectors on each facet and pack them as images for training. Li *et al.* [2016] introduce a probing filter that can efficiently extract features and work for a higher resolution like $64^3$. However, this approach cannot extract fine structures of shapes; therefore, it is not suitable for tasks like shape segmentation. Qi *et al.* [2017] propose a neural network based on an unordered point cloud that can achieve good performance on shape classification and segmentation.

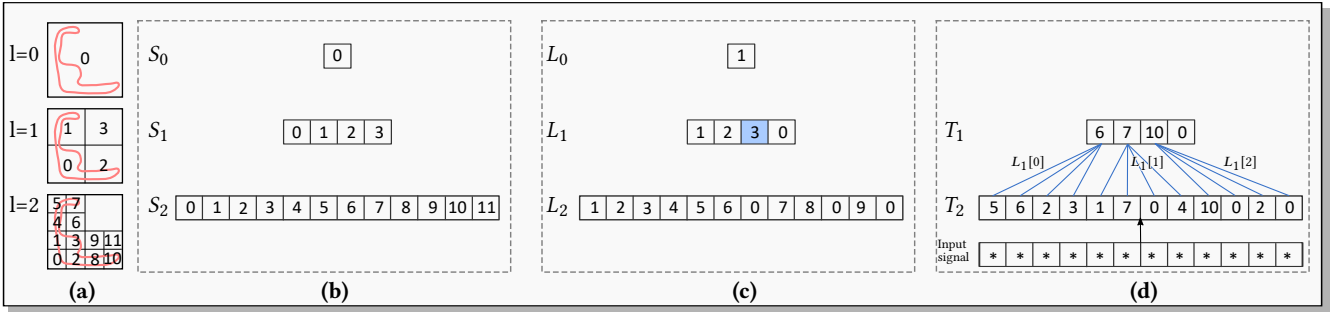

Fig. 2. A 2D quadtree illustration of our octree structure. (a) Given an input 2D shape marked in red, we construct a 2-depth quadtree. The squares that are not occupied by the shape are empty nodes. The numbers inside quad nodes are their shuffled keys. (b) The shuffle key vectors $S_l$, $l = 0 \ldots 2$, each of which stores the shuffled keys of the quad nodes at the $l$-th depth. (c) The label vectors $L_l$, $l = 0 \ldots 2$, each of which stores the labels of the quad nodes at the $l$-th depth. For empty nodes, their labels are set to zero. For non-empty nodes, the label $p$ of a node indicates it is the $p$-th non-empty node at this depth. The label buffer is used to find the correspondence from the parent node to its child nodes. (d) The results of CNN convolutions over the input signal are stored in a feature map $T_2$. When $T_2$ is down-sampled by a CNN operation, such as pooling, the downsampled results are assigned to the first, the second and the third entries of $T_1$. The label vector $L_1$ at the first depth is used to find the correspondence between the nodes at two depths.

*GPU-based octree structures.* The octree structure [Meagher 1982] is widely used in computer graphics for various tasks including rendering, modeling and collision detection. Zhou *et al.* [2011] propose a GPU-based octree construction method and use the constructed octree for reconstructing surfaces on the GPU. Different from the octree structure that is optimized for surface reconstruction, the octree structure designed in our method is optimized for CNN training and evaluation. For this purpose, we discard the pointers from parent octants to children octants and introduce a label array for finding correspondence of octants at different depths for downsampling. Instead of computing the neighborhood of a single octant, we construct the neighborhoods for eight children nodes of a parent node for fast convolution computation.

## 3 3D CNNS ON OCTREE BASED 3D SHAPES

Given an oriented 3D model (e.g. an oriented triangle mesh or a point cloud with oriented normals), our method first constructs the octree of the input 3D model and packs the information needed for CNN operations in the octree (Section 3.1). With the help of this octree structure, all the CNN operations can be efficiently executed on the GPU (Section 3.2).

### 3.1 Octree for 3D CNNs

*Octree construction.* To construct an octree for an input 3D model, we first uniformly scale the 3D shape into an axis-aligned unit 3D bounding cube and then recursively subdivide the bounding cube of the 3D shape in breadth-first order. In each step, we traverse all non-empty octants occupied by the 3D shape boundary at the current depth $l$ and subdivide each of them to eight child octants at the next depth $l + 1$. We repeat this process until the pre-defined octree depth $d$ is reached. Figure 2(a) illustrates a 2-depth quadtree constructed for a 2D shape (marked in red).

After the octree is constructed, we collect a set of properties required for CNN operations at each octant and store their values in the octree. Specifically, we compute a *shuffle key* [Wilhelms and Van Gelder 1992] and a *label* for each octant in the octree. Meanwhile, our method extracts the *input signal* of the CNN from the 3D shape stored in the finest leaf nodes and records the resulting *CNN features* at each octant. As shown in Figure 2, we organize the data stored in the octree in a depth order. At each depth, we sort the octants according to the ascending order of their shuffle keys and pack their property values into a set of 1D property vectors. All the property vectors share the same index, and the length of the vectors is the number of octants at the current depth. In the following, we describe the definition and implementation details of each property defined in our octree structure.

*Shuffle key.* The shuffle key of an octant $O$ at depth $l$ encodes its position in 3D space with an unique 3 $l$-bit string key$(O) := x_1y_1z_1x_2y_2z_2 \ldots x_ly_lz_l$ [Wilhelms and Van Gelder 1992], where each three-bit group $x_i, y_i, z_i \in \{0, 1\}$ defines its relative position in the 3D cube of its parent octant. The integer coordinates $(x, y, z)$ of an octant $O$ are determined by $x = (x_1x_2 \ldots x_l)$, $y = (y_1y_2 \ldots y_l)$, $z = (z_1z_2 \ldots z_l)$. We sort all the octants by their shuffle keys according to the ascending order and store the sorted shuffle keys of all octants at the $l$-th depth in a shuffle key vector $S_l$, which is used later for constructing the neighborhood of an octant for 3D convolution. In our implementation, each shuffle key is stored in a 32 bit integer. Figure 2(a) illustrates the shuffle keys of all quadtree nodes and Figure 2(b) demonstrates the corresponding shuffle key array at each depth.

*Label.* In CNN computations, the pooling operation is frequently used to downsample the features computed at the $l$-th depth octants to their parent octants at the $(l − 1)$-th depth. As a result, we need to quickly find the parent-child relationship of the octants at the adjacent depths. To this end, we assign a label $p$ for a non-empty octant at the $l$-th depth, which indicates that it is the $p$-th non-empty octant in the sorted octant list of the $l$-th depth. For empty octants, we simply set its label as zero. The labels of all octants at the $l$-th depth are stored in a label vector $L_l$. Figure 2(c) illustrates the label vector for each depth of a quadtree. For the non-empty quad node marked in blue, its label is 3 ($L_1[2] = 3$) because it is the third non-empty node at the first depth, while the node with the shuffle key 0 is the first one and the node with the shuffle key 1 is the second one.

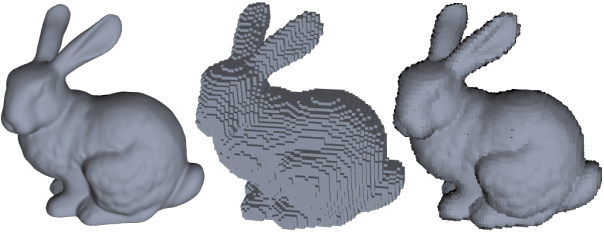

Fig. 3. Left: the original 3D shape. Middle: the voxelized 3D shape. Right: the octree representation with normals sampled at the finest leaf octants.

Given a non-empty node with index $j$ at the $l$-th depth, we compute the index $k$ of its first child octant at the $(l + 1)$-th depth by $k = 8 \times (L_l[j] - 1)$. This is based on two observations. First, only the non-empty octants at the $l$-th depth are subdivided. Second, since we sort the octants according to the ascending order of their shuffle keys, the eight children of an octant are sequentially stored. Moreover, the child octants and their non-empty parents follow the same order in their own property vectors. As shown in Figure 2(c), the last four nodes at the second depth are created by the third non-empty node at the first depth (marked in blue).

*Input signal.* We use the averaged normal vectors computed at the finest leaf octants as the input signal of the CNN. For empty leaf octants, we simply assign a zero vector as the input signal. For non-empty leaf octants, we sample the 3D shape surface embedded in the leaf octant with a set of points and average the normals of all sampled points as the input signal at this leaf octant. We store the input signals of all leaf octants into an input signal vector. The size of the vector is the number of the finest leaf octants in the octree.

Compared to the binary indicator function used in many voxel-based CNN methods [Maturana and Scherer 2015; Wu et al. 2015], the normal signal is very sparse, i.e. only non-zero on the surface, and the averaged normals sampled in the finest leaf octants better represent the orientation of the local 3D shapes and provide more faithful 3D shape information to the CNN. Figure 3 compares a voxelized 3D model and an octree representation of the same 3D shape rendered by oriented disks sampled at leaf octants, where the size of the leaf octant is the same as the voxel size. As shown in the figure, the octree representation (on the right) is more faithful to the ground truth 3D shape (on the left) than the voxel based representation (in the middle).

*CNN features.* For each 3D convolution kernel defined at the $l$-th depth, we record the convolution results on all the octants at the $l$-th depth in a feature map vector $T_l$.

*Mini-batch of 3D models.* For 3D objects in a mini-batch used in the CNN training, their octrees are not the same. To support efficient CNN training on the GPU, we merge these octrees into one super-octree. For each octree depth $l$, we concatenate the property vectors

($S_l$, $L_l$ and $T_l$) of all 3D objects to $S_l^*$, $L_l^*$ and $T_l^*$ of the super-octree. After that, we update the shuffle keys in $S_l^*$ by using the highest 8 bits of each shuffle key to store the object index. We also update the label vector $L_l^*$ in the super-octree to represent the index of each non-empty octant in the whole super-octree. After that, the super-octree can be directly used in the CNN training.

### 3.2 CNN operations on the Octree

The most common operations in a CNN are convolution, pooling, and the inverse deconvolution and unpooling. With the help of our octree data structure, all these CNN operations can be efficiently implemented on the GPU.

*3D Convolution.* For applying the convolution operator to an octant, one needs to pick its neighboring octants at the same octree depth. To compute the convolution efficiently, we write the convolution operator $\Phi_c$ in the unrolled form:

$$\Phi_c(O) = \sum_n \sum_i \sum_j \sum_k W_{ijk}^{(n)} \cdot T^{(n)}(O_{ijk}).$$

Here $O_{ijk}$ represents a neighboring octant of $O$ and $T(\cdot)$ represents the feature vector associated with $O_{ijk}$. And $T^{(n)}(\cdot)$ represents the $n$-th channel of the feature vector, and $W_{ijk}^{(n)}$ are the weights of the convolution operation. If $O_{ijk}$ does not exist in the octree, $T(O_{ijk})$ is set to the zero vector. In this form, the convolution operation can be converted to a matrix product [Chellapilla et al. 2006; Jia et al. 2014] and computed efficiently on the GPU.

The convolution operator with kernel size $K$ requires the access of $K^3 - 1$ neighboring octants of an octant. A possible solution is to pre-compute and store neighboring information. However, a CNN is normally trained on a batch of shapes. When $K$ is very large, this will cause a significant amount of I/O processing and a large memory footprint. We build a hash table: $\mathcal{H} : key(O) \mapsto index(O)$ to facilitate the search, where index($O$) records the position of $O$ in $S$. Because the amortized time complexity of a hash table is constant, this choice can be regarded as a balance of computation cost and memory cost for the CNN. Given the shuffled key stored in the vector $S_l$, the integer coordinates ($x, y, z$) of the octant can be restored, then the coordinates of the neighboring octants can be computed easily in constant time, as can their corresponding shuffled keys. Given the shuffled keys of neighboring octants, the hash table is efficiently searched in parallel to get their indices, according to which the neighboring data information is gathered and then the convolution operation can be applied.

If the stride of convolution is 1, the above operation is applied to all existing octants at the current octree depth. Naïvely, for each octant the hash table will be searched $K^3 - 1$ times. However, the neighborhood of eight sibling octants under the same parent have a lot of overlap, and there are only $(K+1)^3$ individual octants including the eight octants. The neighborhood search can be further sped up by just searching the $(K + 1)^3 - 8$ neighboring octants for each of the eight sibling octants. Concretely, if the kernel size is 3, then this optimization can accelerate the neighboring search operation by more than 2 times. Figure 4 illustrates this efficient neighborhood search on a 2D quadtree.

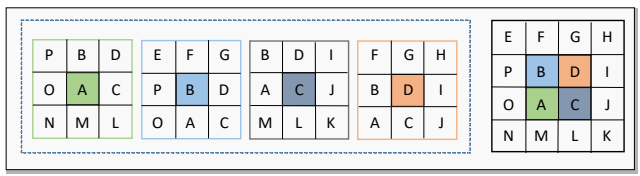

Fig. 4. Neighbor access in convolution in 2D. A naïve implementation would pick 36 neighbors for the 4 sibling nodes (A, B, C, D) as shown on the left. Since the neighborhood of 4 sibling nodes under the same parent have a lot of overlap, we only access the union of these neighbors whose total number is 16 as shown on the right.

It is also very easy to perform a convolution with a stride of 2. Specifically, for the 8 sibling octants under the same parent, the convolution can be applied to the first sibling octant while ignoring the other siblings, which is equivalent to down-sampling the resolution of the feature map by a factor of 2. As for the convolution with a stride of $2^r (r > 1)$, the operation can be applied to the first octant belonging to each sub-tree of height $r$, then the feature map will be downsampled by a factor of $2^r (r > 1)$. Because of the special hierarchical structure of an octree, the stride of convolution is constrained to be an integer power of 2. However, convolution with an arbitrary stride is uncommon in the CNN literature. According to our experiments, the performance of our method will not be harmed by this stride limitation.

When performing the convolution operation with a stride larger than 1, down-sampling occurs and the length of the data vector $T$ is shortened. The data flows from bottom to top. Then information stored in $L_l$ can be used to get the correspondence. Take the example shown in Figure 2 as an illustration, the initial length of $T_2$ is 12. When down-sampling occurs the length of the vector will be 3. But there are 4 octants at the first depth of the octree, and the length of $T_1$ should be 4. Combining the information stored in $L_1$ and the down-sampled vector, $T_1$ can be renewed easily.

*Pooling.* The main functionality of pooling is to progressively condense the spatial size of the representation. The pooling layer operates independently on every channel of the feature map and resizes it spatially. The most common form is the max-pooling layer with filters of kernel size 2 applied with a stride of 2. It is very convenient to apply pooling on our octree structure. Since every 8 sibling octants under the same parent are stored consecutively, applying the max-pooling operator on an octree reduces to picking out the max elements from every 8 contiguous elements, which can be implemented efficiently on GPU devices. Then the resolution of the feature map is down-sampled by a factor of 2, and the information from the parent octants can be used to guide further operations.

Since pooling can be regarded as a special kind of convolution in practice, we follow the approach presented before for general pooling operations with other kernel sizes or stride sizes, i.e. finding the corresponding neighboring octants and applying the specified operation, such as max-pooling or average-pooling.

*Unpooling.* The unpooling operation is the reverse operation of pooling and performs up-sampling, which is widely used in CNN visualization [Zeiler and Fergus 2014] and image segmentation [Noh et al. 2015]. The max-unpooling operation is often utilized together with the max-pooling operation. After applying the max-pooling operation, the locations of the maxima within each pooling region can be recorded in a set of switch variables stored in a continuous array. The corresponding max-unpooling operation makes use of these switches to place the signal of the current feature map into appropriate locations of the up-sampled feature map. Figure 5 shows how max-unpooling works on a quadtree. Thanks to the contiguous storage of octants, we can reuse the efficient unpooling implementation developed for image-based CNNs.

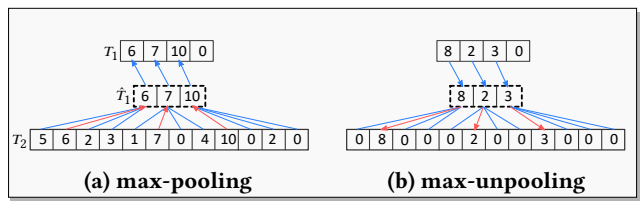

Fig. 5. An example of max-pooling/unpooling on a quadtree structure of Fig. 2. Since every 4 sibling nodes under the same parent are stored contiguously, applying the max-pooling operator on a quadtree reduces to picking out the max element for each of 4 contiguous elements in an array. After down-sampling, the intermediate results stored in a temporary array $\hat{T}_1$ as shown by the dashed box. Then the label vector $L_1$ is used to construct $T_1$, i.e. assigning $\hat{T}_1[L_1[i] - 1]$ to $T_1[i]$ when $L_1[i] > 0$, and 0 otherwise. The unpooling operation is the reverse operation of pooling. The red arrows represent the switch variables.

*Deconvolution.* The deconvolution operator, also called *transposed convolution* and *backwards convolution* [Long et al. 2015; Zeiler and Fergus 2014], can be used to enlarge and densify the activation map, which can be implemented by just reversing the forward and backward passes of convolution. Based on the convolution on octrees presented previously, the deconvolution operation can be implemented accordingly.

*Remark.* Different from full-voxel-based CNNs that perform CNN operations over all the space including empty regions, CNN operations in our method are applied to octants only. That is to say, *where there is an octant, there is CNN computation*. Our understanding here is that propagating information to empty regions and exchanging information via empty regions are not necessary and would require more memory and computations. By restricting the information propagation in the octree, the shape information can be exchanged more effectively along the shape. Although there is not a theoretical proof on this point, we demonstrate the advantages of our method in Section 5.

By restricting the data storage and CNN computations in octants, the memory and computation cost of the octree based CNN is $O(n^2)$, where $n$ is the voxel resolution in each dimension at the finest level. On the contrary, the memory and computational cost of a full-voxel based solution is $O(n^3)$. Furthermore, since all the data is stored contiguously in memory, O-CNN shares the same high performance GPU computations with 2D and 3D CNNs defined on the regular grids. The hash table and neighbor information which are needed in the $k$-th iteration during the training can be pre-computed in the $(k-1)$-th iteration in a separate thread, which incurs no computation time latency. A detailed evaluation and comparison are performed in Section 5.1.

## 4 NETWORK STRUCTURE

The network structure of CNNs has evolved rapidly in recent years. Deeper and wider networks have shown their superiority in accomplishing many tasks. Existing 3D CNNs have used different networks to enhance their capabilities. However, this makes it hard to distinguish where the main benefit of their approach comes from. To clearly demonstrate the advantages of our octree-based representation, we design a simple network by following the concept of LeNet [Lecun et al. 1998].

*O-CNN*. Our O-CNN is simple: we repeatedly apply convolution and pooling on the octree data structure from bottom to top. We use the ReLU function ($f : x \in \mathbb{R} \mapsto \max(0, x)$) to activate the output and use batch normalization (BN) to reduce the internal-covariate-shift [Ioffe and Szegedy 2015]. We call the operation sequence "convolution + BN + ReLU + pooling" a *basic unit* and denote it by $U_l$ if the convolution is applied to the $l$-th depth octants. The number of channels of the feature map for $U_l$ is set to $2^{\max(1, 9-l)}$ and the convolution kernel size is 3. Our O-CNN is defined by the following form:

$$\text{input} \rightarrow U_d \rightarrow U_{d-1} \rightarrow \cdots \rightarrow U_2$$

and we call it O-CNN(d). To align all features from different octree structures, we enforce all the 2nd-depth octants to exist and use zero vector padding on the empty octants at the 2nd depth.

*O-CNN for shape analysis*. In our work, we apply our O-CNN to three shape analysis tasks: object classification, shape retrieval, and shape part segmentation.

For object classification, we add two fully connected (FC) layers, a softmax layer, and two Dropout layers [Srivastava et al. 2014] after O-CNN(d), i.e.

$$\text{O-CNN}(d) \rightarrow \text{Dropout} \rightarrow \text{FC}(128) \rightarrow \text{Dropout} \rightarrow$$
$$\text{FC}(N_c) \rightarrow \text{softmax} \rightarrow \text{output}.$$

Here 128 is the number of neurons in FC and $N_c$ is the number of classification categories. Dropout is used to avoid overfitting.

For shape retrieval, we use the output from the object classification as the key to search for the most similar shapes to the query.

For shape part segmentation, we adopt the state-of-the-art image semantic segmentation network *DeconvNet* [Noh et al. 2015], which cascades a deconvolution network after a convolution network for dense predictions. The convolution network is set as our O-CNN(d). The deconvolution network is the mirror of O-CNN(d) where the convolution and pooling operators are replaced by deconvolution and unpooling operators. We define "unpooling + deconvolution + BN + ReLU" as a basic unit and denote it by $DU_l$ if the unpooling is applied to the $l$-depth octants. The network structure for shape segmentation is

$$\text{O-CNN}(d) \rightarrow DU_2 \rightarrow DU_3 \rightarrow \cdots \rightarrow DU_d.$$

The details of O-CNN for the above tasks and the experiments are presented in the next section.

## 5 EXPERIMENTS AND DISCUSSION

For demonstrating the efficiency and efficacy of our O-CNN, we conduct three shape analysis tasks on a desktop machine with an Intel Core I7-6900K CPU (3.2 GHz) and a GeForce 1080 GPU (8GB memory). Our GPU implementation of O-CNN uses the Caffe framework [Jia et al. 2014] and is available at http://wang-ps.github.io/O-CNN.

*Training details*. We optimize the O-CNNs by stochastic gradient descent (SGD) with a momentum of 0.9, a weight decay of 0.0005, and a batch size of 32. The dropout ratio is 0.5. The initial learning rate is 0.1, and decreased by a factor of 10 after every 10 epochs. The optimization stops after about 40 epochs. The hyper-parameters of the network are fixed in the shape classification and retrieval experiments. They are fine-tuned in object segmentation for the categories with a small number of shapes.

*Octree data preparation*. For the tasks we perform, the 3D training datasets are mainly from ModelNet40 [Wu et al. 2015] and ShapeNetCore55 [Chang et al. 2015] which contain a large number of triangle meshes with various shapes. We find that many meshes in ModelNet contain a lot of artifacts: flipped normals, non-manifold structures, and overlapped triangles. Thus to build the octree data structure with correct normal information, we first use the ray shooting algorithm to sample dense points with oriented normals from the shapes. Specifically, we place 14 virtual cameras on the face centers of the truncated bounding cube of the object, uniformly shoot 16k parallel rays towards the object from each direction, calculate the intersections of the rays and the surface, and orient the normals of the surface points towards the camera. The points on the invisible part of the shapes are discarded. We then build an octree structure on the point cloud and compute the average normal vectors of the points inside the leaf octants. The octree structures of all the shapes are stored in a database and saved on a hard disk.

### 5.1 Object classification

The goal of object classification is to assign category information to every object, which is an essential and fundamental task in understanding 3D shapes.

*Dataset*. We use the ModelNet40 dataset [Wu et al. 2015] for training and testing, which contains 12,311 CAD models from 40 categories, and is well annotated with multi-class labels. The training and testing sets are available in the dataset, in which 9,843 models are used for training, and 2,468 models for testing. The upright orientation of the models in the dataset is known. We augment the dataset by rotating each model along the upright direction uniformly to generate 12 poses for each model.

*Training details*. Using the augmented dataset, we train our O-CNN(d) as shown in Section 4. To observe the behavior of O-CNN under different resolutions, we train six networks: O-CNN(3), O-CNN(4), O-CNN(5), O-CNN(6), O-CNN(7), O-CNN(8), i.e. the resolutions of leaf octants are $8^3$, $16^3$, $32^3$, $64^3$, $128^3$, $256^3$, respectively. The loss function is modeled as the cross-entropy, which is commonly used for classification.

| Network | without voting | with voting |
|---|---|---|
| VoxNet ($32^3$) | 82.0% | 83.0% |
| Geometry image | 83.9% | - |
| SubVolSup ($32^3$) | 87.2% | 89.2% |
| FPNN ($64^3$) | 87.5% | - |
| FPNN+normal($64^3$) | 88.4% | - |
| PointNet | 89.2% | - |
| VRN ($32^3$) | 89.0% | **91.3%** |
| O-CNN(3) | 85.5% | 87.1% |
| O-CNN(4) | 88.3% | 89.3% |
| O-CNN(5) | 89.6% | 90.4% |
| O-CNN(6) | **89.9%** | 90.6% |
| O-CNN(7) | 89.5% | 90.1% |
| O-CNN(8) | 89.6% | 90.2% |

Table 1. Object classification results on ModelNet40 dataset. The number shown in the table is the accuracy of object recognition. The second and third columns show the results of the network with and without voting. Numbers in parentheses are the resolutions of voxels. A number in boldface emphasizes the best result.

*Orientation pooling.* Since each model is rotated to 12 poses, in the testing phase the activations of the output layer for each pose can be pooled together to increase the accuracy of predictions. In this case, the orientation pooling is reduced to the voting approach, which has been adopted by [Maturana and Scherer 2015]. More effectively, one can also pool the activations of the last convolution layers and then fine-tune the last two FC layers. This strategy has been adopted by [Qi et al. 2016; Su et al. 2015]. But the disadvantage of this approach is that it requires the training of one more neural network. For simplicity, we choose the voting strategy for classification. In Table 1, we provide results with and without voting.

*Comparisons and discussion.* We did a comparison on classification accuracy with state-of-the-art 3D CNNs: VoxNet [Maturana and Scherer 2015], SubVolSup [Qi et al. 2016], FPNN [Li et al. 2016], PointNet [Qi et al. 2017], Geometry image [Sinha et al. 2016], and VRN [Brock et al. 2016]. For fair comparison, we only consider the performance of a single CNN, and omit results from an ensemble of CNNs.

From Table 1, we find that our O-CNN has a significant advantage over VoxNet and FPNN. Our O-CNN(4) whose resolution is $16^3$ is already better than FPNN($64^3$) and slightly worse than FPNN+normal($64^3$) which utilizes both the distance field and the normal field information. Compared with non voxel-based methods, O-CNN(4) is only worse than PointNet. When we increase the resolution, O-CNN(5) beats all other methods. With the voting strategy, O-CNN is only worse than VRN which uses 24 rotation copies for training and voting.

It is also interesting to see that O-CNN(3) already has good accuracy, which is above 85%. This fact is consistent with human recognition: people can recognize the type of 3D shape easily from far away. The result indicates that our octree with shape normal information is very informative.

We observe that the accuracy of our O-CNN increases gradually with the resolution. But there are small accuracy drops when the resolution exceeds $64^3$. This is probably because the ModelNet40 dataset is still not big enough and when the networks go deeper there are some overfitting risks during training.

| data structure | input signal | without voting |
|---|---|---|
| full voxel | binary | 87.9% |
| full voxel | normal | 88.7% |
| octree | binary | 87.3% |
| octree | normal | **89.6%** |

Table 2. Representation comparison on the ModelNet40 dataset with the same network architecture. The representation resolution is $32^3$. The number shown in the table is the accuracy of classification without voting.

*Representation comparison.* To further verify the superiority of using the octree structure with the normal signals, we have done experiments with the same network architecture as O-CNN(5) while using different input representations. The representation variation includes: (1) full voxel with the binary signal; (2) full voxel with the normal signal with zero vectors in empty voxels; (3) octree with the normal signal, i.e. the representation for our O-CNN; (4) octree with the binary signal, i.e. replacing the normal signal with occupying bits. The second variation can be regarded a generalized version of octree with the normal signal. For CNNs with the full voxel representation in our tests, we adapt the implementation of VoxNet [Maturana and Scherer 2015] for our network architecture. We train and test the network on the ModelNet40 dataset, with the results summarized in Table 2.

Note that the normal signal helps to achieve better performance on both octree and full voxel structure, which verifies our claim in Section 3.1 that the normal signal preserves more information of the original shape and is superior over the binary signal.

Moreover, the octree with the normal signal provides the highest level of accuracy among all methods, while the full voxel with the normal signal that performs the computation everywhere does not yield a better result. This indicates that restricting the computation on the octants only is a reasonable strategy that results in the good performance of the O-CNN. We leave a rigorous theoretical analysis for future studies.

Finally, we found that the octree with the binary signal has the worse performance than full voxel structure with binary signal. This is because the indicator function that represents the original shape is defined in a volume, while our octree is built from the point cloud. After replacing the normal signal as the occupying bits, it is equivalent to discarding the inside portion of the indicator function, which causes information loss compared with the full voxel representation and makes it hard to distinguish the inside and outside of the object.

*Comparisons of memory and computation efficiency.* We compare the memory and computational costs of our O-CNN with the costs of full-voxel-based CNNs. Similar to the testing done in *representation comparison*, we use our network structure for full-voxel-based CNNs. Note that in our network structure, the channel number of the feature maps at each depth used in O-CNNs decreases by a factor of 2 as the octree depth increases, the memory cost of O-CNN(d) can be reduced to be $O(n)$, which enables O-CNN to analyze the 3D shapes in high resolutions.

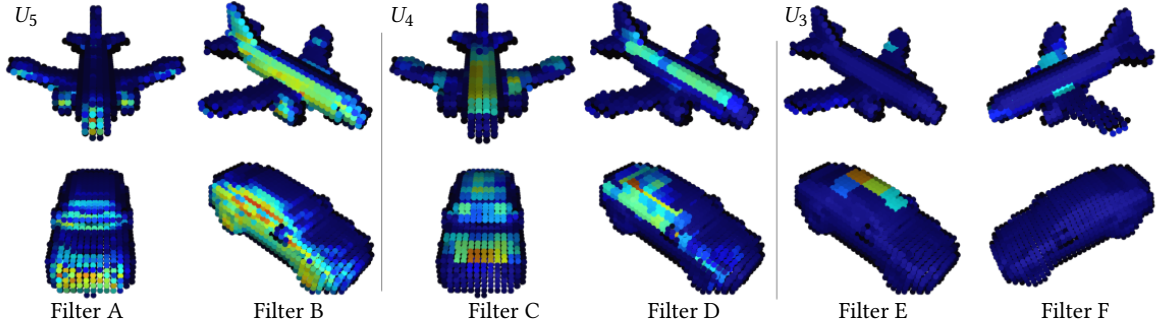

Fig. 6. CNN visualization. The responses of some convolutional filters at different levels on two models are rendered. Red represents a large response and blue a low response.

| Method | $16^3$ | $32^3$ | $64^3$ | $128^3$ | $256^3$ |
|---|---|---|---|---|---|
| O-CNN | 0.32GB | 0.58GB | 1.1GB | 2.7GB | 6.4GB |
| full voxel+binary | 0.23GB | 0.71GB | 3.7GB | Out of memory | Out of memory |
| full voxel+normal | 0.27GB | 1.20GB | 4.3GB | Out of memory | Out of memory |

Table 3. Comparisons on GPU-memory consumption. The batch size is 32.

| Method | $16^3$ | $32^3$ | $64^3$ | $128^3$ | $256^3$ |
|---|---|---|---|---|---|
| O-CNN | 17ms | 33ms | 90ms | 327ms | 1265ms |
| full voxel+binary | 59ms | 425ms | 1648ms | - | - |
| full voxel+normal | 75ms | 510ms | 4654ms | - | - |

Table 4. Timings of one backward and forward operation in milliseconds. The batch size is 32.

We run 1000 forward-backward iterations, including all CPU-GPU communications, calculate the average time per iteration, and record the peak GPU memory consumption. The statistics of memory and time cost are summarized in Tables 3 and 4. It is clear that the O-CNN runs much faster under all resolutions and occupies less memory when the resolution is greater than $16^3$ (i.e. $d \leq 4$).

*CNN visualization.* With respect to CNNs for image understanding, it is known that the output of the learned convolution filters [Goodfellow et al. 2016] is activated when important image features appear. We also observe this phenomenon on our O-CNN and it helps to better understand O-CNN.

In Figure 6, we illustrate some filters in $U_5$, $U_4$, $U_3$ of O-CNN(5) by color-coding the responses to the input shape on the corresponding octants. We find that the filters in $U_5$ capture low-level geometry features of the shape, while the filters in $U_4$ and $U_3$ capture high-level shape features. In $U_5$, filter A tends to capture forward planar regions, filter B prefers large round regions. In $U_4$, filters C and D capture more global shape features. In $U_3$, filters E and F are more sensitive to the shape category and they have little response when the category of the shape does not fit.

## 5.2 Shape retrieval

Nowadays 3D shapes are widely available, to manage and analyze them, a 3D shape retrieval method is essential. We test our O-CNN on the large-scale dataset ShapeNet Core55 and compare with the state-of-the-art methods in the 3D shape retrieval contest – SHREC16 [Savva et al. 2016].

*Dataset.* The ShapeNet Core55 dataset contains a total of 51190 3D models with 55 categories and 204 subcategories. The models are normalized to a unit length cube and have a consistent upright orientation. 70% of the dataset is used for training, 10% for validation, and 20% for testing. We use the same method as object classification to perform data augmentation.

*Retrieval.* The key to shape retrieval is to generate a compact and informative feature for each shape, with which the most similar shape can be retrieved. We train an O-CNN as the feature extractor, and the network structure is the same as the one used for classification. In the training stage, the cross-entropy loss function is minimized with only the category information. The subcategory information in the dataset is discarded for simplicity. The O-CNN output is the category probability of the input shape, which is used as the feature vector of each shape. Since each object is rotated to 12 poses, correspondingly there are 12 feature vectors, with which the orientation pooling presented previously is used to generate one feature vector for each shape. For each query shape, the label can be predicted from this feature vector. The retrieval set of a query shape is constructed by collecting all shapes that have the same label, and then sorting them according to the feature vector distance between the query shape and the retrieved shape. Fig. 7 shows top-5 retrieval results of three models by O-CNN(6).

*Performance comparison.* For each query in the test set, a retrieval list is returned. Five metrics are used to evaluate the quality of results: precision, recall, mAP, F-score, and NDCG. The precision at an entry is the fraction of retrieved instances that are relevant up to this entry, while the recall is the fraction of relevant instances up to this entry. Roughly, the recall increases along with the length of the retrieval list, while precision decreases. mAP is the mean average precision and F-score is defined as the harmonic mean of precision and recall, which can be regarded as a summary of precision and recall. NDCG measures ranking quality, and the subcategory similarity between shapes is considered when computing this metric. The average performance across all query shapes on these four metrics are calculated with the evaluation software officially provided by [Savva et al. 2016], and summarized in Table 5. The precision recall curve is shown in Figure 8.

We compare our O-CNN with five state-of-the-art methods in SHREC16 which are all based on multi-view based CNNs [Bai et al.

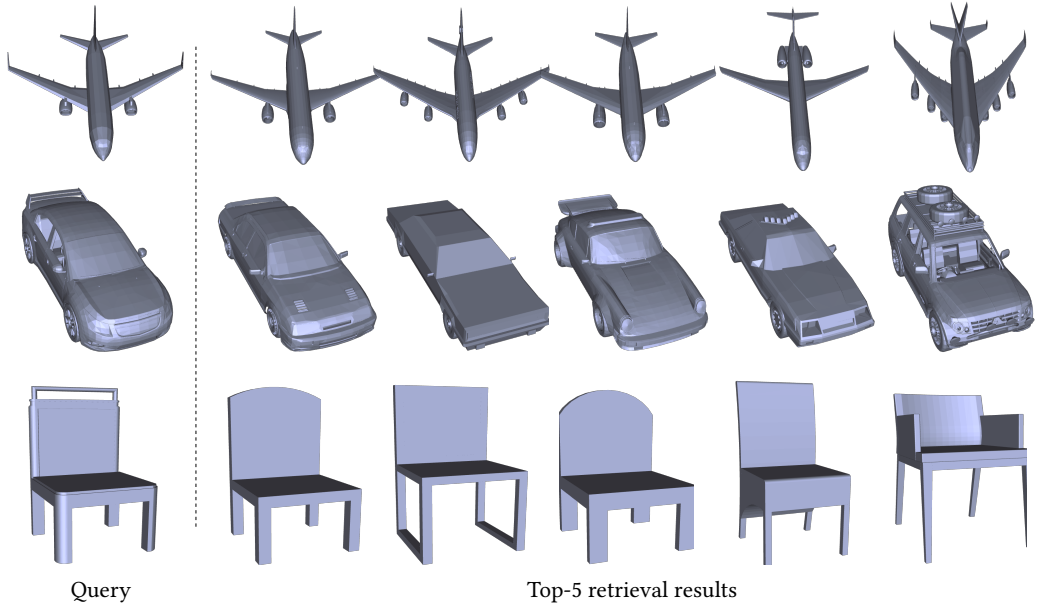

Fig. 7. Top-5 retrieval results of O-CNN(6) on three models.

| Method | P@N | R@N | F1@N | mAP | NDCG@N |
|---|---|---|---|---|---|
| Tatsuma_DB-FMCD-FUL-LCDR | 0.427 | 0.689 | 0.472 | 0.728 | 0.875 |
| Wang_CCMLT | 0.718 | 0.350 | 0.391 | 0.823 | 0.886 |
| Li_ViewAggregation | 0.508 | **0.868** | 0.582 | 0.829 | 0.904 |
| Bai_GIFT | 0.706 | 0.695 | 0.689 | 0.825 | 0.896 |
| Su_MVCNN | 0.770 | 0.770 | 0.764 | 0.873 | 0.899 |
| O-CNN(5) | 0.768 | 0.769 | 0.763 | 0.871 | 0.904 |
| O-CNN(6) | **0.778** | 0.782 | **0.776** | **0.875** | **0.905** |

Table 5. Retrieval results. The upper five methods are from the teams that submitted results to SHREC16.

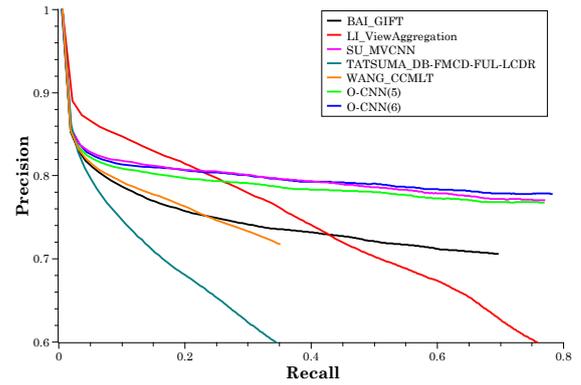

Fig. 8. Precision and recall curves.

2016; Savva et al. 2016; Su et al. 2015]. As we can see from Table 5, our O-CNN(5) is comparable to state-of-the-art results, and O-CNN(6) yields the best results among all tested methods. Moreover, though the subcategory information is discarded when training our network, we also get the best score on NDCG, which shows that with our octree representation the learned feature is very discriminative and can distinguish similar shapes very well.

### 5.3 Object part segmentation

Given a 3D object represented by a point cloud or a triangle mesh, the goal of part segmentation is to assign part category information to each point or triangle face. Compared with object classification, part segmentation is more challenging since the prediction is fine-grained and dense.

*Dataset.* We conduct an experiment on a large-scale shape part annotation dataset introduced by [Yi et al. 2016], which augments a subset of the ShapeNet models with semantic part annotations. The dataset contains 16 categories of shapes, with 2 to 6 parts per category. In total there are 16,881 models with part annotations. However, the models in this dataset are represented as sparse point clouds, with only about 3k points for each model, and the point normals are missing. We align the point cloud with the corresponding 3D mesh, and project the point back to the triangle faces. Then we assign the normal of the triangle face to the point, and condense the point cloud by uniformly re-sampling the triangle faces. Based on this pre-processed point cloud, the octree structure is built. Similar to other tasks, every model has 12 copies rotated around the upright axis. For comparison with [Qi et al. 2017; Yi et al. 2017], we use the same training/test split.

*Training.* Compared with the dataset used in the retrieval and classification task, the dataset for segmentation is still limited, so training this network from scratch is challenging. Instead, we reuse the weights trained by the retrieval task. Specifically, in the training stage, the convolution part is initialized with the weight trained on ShapeNet and fixed during optimization, while the weight of the deconvolution part is randomly initialized and then evolves according to the optimization process.

| | mean | plane | bag | cap | car | chair | e.ph. | guitar | knife | lamp | laptop | motor | mug | pistol | rocket | skate | table |
|---|---|---|---|---|---|---|---|---|---|---|---|---|---|---|---|---|---|
| # shapes | | 2690 | 76 | 55 | 898 | 3758 | 69 | 787 | 392 | 1547 | 451 | 202 | 184 | 283 | 66 | 152 | 5271 |
| [Yi et al. 2016] | 81.4 | 81.0 | 78.4 | 77.7 | 75.7 | 87.6 | 61.9 | 92.0 | 85.4 | 82.5 | **95.7** | **70.6** | 91.9 | **85.9** | 53.1 | 69.8 | 75.3 |
| PointNet [Qi et al. 2017] | 83.7 | 83.4 | 78.7 | 82.5 | 74.9 | 89.6 | 73.0 | 91.5 | 85.9 | 80.8 | 95.3 | 65.2 | 93.0 | 81.2 | 57.9 | 72.8 | 80.6 |
| SpecCNN [Yi et al. 2017] | 84.7 | 81.6 | 81.7 | 81.9 | 75.2 | 90.2 | 74.9 | 93.0 | 86.1 | **84.7** | 95.6 | 66.7 | 92.7 | 81.6 | **60.6** | **82.9** | 82.1 |
| O-CNN(5) | 85.2 | 84.2 | 86.9 | 84.6 | 74.1 | 90.8 | 81.4 | 91.3 | 87.0 | 82.5 | 94.9 | 59.0 | 94.9 | 79.7 | 55.2 | 69.4 | 84.2 |
| O-CNN(6) | **85.9** | **85.5** | **87.1** | **84.7** | **77.0** | **91.1** | **85.1** | 91.9 | **87.4** | 83.3 | 95.4 | 56.9 | **96.2** | 81.6 | 53.5 | 74.1 | **84.4** |

Table 6. Object part segmentation results.

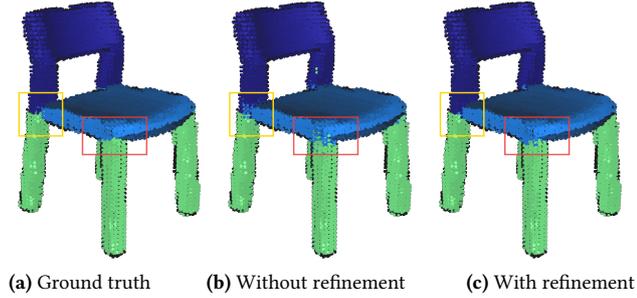

**(a)** Ground truth  **(b)** Without refinement  **(c)** With refinement

Fig. 9. The effect of CRF refinement. (a) The ground truth part segmentation of a chair. (b) The result of O-CNN(6) without CRF refinement. As highlighted by the boxes, the boundaries of segmentation results are noisy and jagged. (c) The result of O-CNN(6) with CRF refinement. The inconsistencies on the segmentation boundaries are greatly reduced.

*CRF refinement.* After training, very promising part category predictions can already be achieved according to the output of the deconvolution network. However, there are still some inconsistencies among adjacent segmented parts as shown in Fig. 9(b) because the deconvolution network makes the prediction for each point separately. We adopt the dense conditional random field (CRF) technique [Krähenbühl and Koltun 2011] to refine results.

For the point cloud $\{\mathbf{p}_i\}_{i=1}^N$ with the corresponding normal $\{\mathbf{n}_i\}_{i=1}^N$, we denote the predicted label as $\{x_i\}_{i=1}^N$. The following energy function is minimized to obtain the refined output:

$$E(x) = \sum_i \phi_u(x_i) + \sum_{i<j} \phi_p(x_i, x_j)$$

where $i$ and $j$ range from 1 to $N$. $\phi_u(x_i)$ is the unary energy, and is defined as $\phi_u(x_i) = -\log(p(x_i))$, which is used to constrain the final output CRF to be similar with our neural network, where $p(x_i)$ is the label probability produced by the neural network. $\phi_p(x_i, x_j)$ is the pairwise energy to incorporate neighbor information to refine the output:

$$\phi_p(x_i, x_j) = \mu(x_i, x_j) \left( \omega_1 W_{\theta_1}(\|\mathbf{p}_i - \mathbf{p}_j\|) + \omega_2 W_{\theta_2}(\|\mathbf{p}_i - \mathbf{p}_j\|) W_{\theta_3}(\|\mathbf{n}_i - \mathbf{n}_j\|) \right)$$

where $W_{\theta_i}$ is the Gaussian function with a standard deviation $\theta_i$. The label compatibility function $\mu(x_i, x_j)$, and hyper-parameters $\omega_i$ and $\theta_i$ are learned on the training set with the implementation provided by [Krähenbühl and Koltun 2013].

With CRF refinement, the inconsistencies between segmented parts can be significantly reduced. Fig. 9(c) shows the refinement result of Fig. 9(b).

*Performance comparison.* The comparison is conducted with a learning-based technique [Yi et al. 2017] that leverages per-point local geometric features and correspondences between shapes, and two recent deep learning based methods [Qi et al. 2017; Yi et al. 2017]. The evaluation metric is the intersection over union (IoU) of the part class. For each shape category, all the part class IoUs are averaged together to obtain the category IoU. Table 6 reports our results. O-CNN(5) and O-CNN(6) perform better or comparable to other methods. O-CNN(6) is also better than O-CNN(5) in many cases and sufficiently demonstrates the benefits of using high resolution.

## 6 CONCLUSION

We propose octree-based convolutional neural networks (O-CNN) that take advantage of the sparseness of the octree representation and the local orientation of the shape to enable compact storage and fast computation, achieving better or comparable performance to existing work. The experiments on three shape analysis tasks demonstrate the efficacy and efficiency of O-CNNs. We expect that O-CNN will stimulate more work on 3D understanding and processing.

In the future, we would like to use our O-CNN to solve more shape analysis and processing challenges, especially on fine-grained tasks where O-CNNs with high resolutions are essential, like shape denoising, shape correspondence, object generation, and scene analysis. There are also many directions for improving O-CNNs, such as the following:

*Adaptive octree.* In our octree construction, we do not consider the geometry change of the shape. Actually, for nearly flat regions, it is fine to use a bigger octant to represent them without subdivision. By constructing the octree adaptively according to the local geometry, it is possible to further improve the computation and memory efficiency of O-CNN.

*General lattices.* In our O-CNN, we organize the 3D data and computation in the octree data structure, which can be regarded as a hierarchical sparse grid lattice. It is possible to build similar hierarchical structures based on other lattices, such as the tetrahedral lattice [Graham 2015] and the permutohedral lattice [Jampani et al. 2016], the latter of which could be used for higher-dimension CNNs. We leave the generalization of these lattice structures for future studies.

*Network structure.* Although the structure of CNNs plays an important role in improving performance, we have not experimented with advanced structures, like deep residual networks or recurrent neural networks. In the future, we would like to integrate our O-CNNs with these advanced network structures to accomplish more challenging tasks.

O-CNN: Octree-based Convolutional Neural Networks for 3D Shape Analysis • 72:11

## ACKNOWLEDGMENTS

We wish to thank the authors of [Chang et al. 2015; Wu et al. 2015] for sharing their 3D model datasets with the public, the authors of [Qi et al. 2017; Yi et al. 2016] for providing their evaluation details, Stephen Lin for proofreading the paper, and the anonymous reviewers for their constructive feedback.

## REFERENCES

Song Bai, Xiang Bai, Zhichao Zhou, Zhaoxiang Zhang, and Longin Jan Latecki. 2016. GIFT: A real-time and scalable 3D shape search engine. In *Computer Vision and Pattern Recognition (CVPR)*.

D. Boscaini, J. Masci, S. Melzi, M. M. Bronstein, U. Castellani, and P. Vandergheynst. 2015. Learning class-specific descriptors for deformable shapes using localized spectral convolutional networks. *Comput. Graph. Forum* 34, 5 (2015), 13–23.

Davide Boscaini, Jonathan Masci, Emanuele Rodolà, and Michael M. Bronstein. 2016. Learning shape correspondence with anisotropic convolutional neural networks. In *Neural Information Processing Systems (NIPS)*.

Andrew Brock, Theodore Lim, J.M. Ritchie, and Nick Weston. 2016. Generative and discriminative voxel modeling with convolutional neural networks. In *3D deep learning workshop (NIPS)*.

M. M. Bronstein, J. Bruna, Y. LeCun, A. Szlam, and P. Vandergheynst. 2017. Geometric deep learning: going beyond Euclidean data. *IEEE Sig. Proc. Magazine* (2017).

Angel X. Chang, Thomas Funkhouser, Leonidas Guibas, Pat Hanrahan, Qixing Huang, Zimo Li, Silvio Savarese, Manolis Savva, Shuran Song, Hao Su, Jianxiong Xiao, Li Yi, and Fisher Yu. 2015. ShapeNet: an information-rich 3D model repository. arXiv:1512.03012 [cs.GR]. (2015).

Kumar Chellapilla, Sidd Puri, and Patrice Simard. 2006. High performance convolutional neural networks for document processing. In *International Conference on Frontiers in Handwriting Recognition (ICFHR)*.

Ian Goodfellow, Yoshua Bengio, and Aaron Courville. 2016. *Deep Learning*. MIT Press.

Ben Graham. 2015. Sparse 3D convolutional neural networks. In *British Machine Vision Conference (BMVC)*.

Kan Guo, Dongqing Zou, and Xiaowu Chen. 2015. 3D mesh labeling via deep convolutional neural networks. *ACM Trans. Graph.* 35, 1 (2015), 3:1–3:12.

Varun Jampani, Martin Kiefel, and Peter V. Gehler. 2016. Learning sparse high dimensional filters: image filtering, dense CRFs and bilateral neural networks. In *Computer Vision and Pattern Recognition (CVPR)*.

Yangqing Jia, Evan Shelhamer, Jeff Donahue, Sergey Karayev, Jonathan Long, Ross Girshick, Sergio Guadarrama, and Trevor Darrell. 2014. Caffe: convolutional architecture for fast feature embedding. In *ACM Multimedia (ACMMM)*. 675–678.

Philipp Krähenbühl and Vladlen Koltun. 2011. Efficient inference in fully connected CRFs with gaussian edge potentials. In *Neural Information Processing Systems (NIPS)*.

Philipp Krähenbühl and Vladlen Koltun. 2013. Parameter learning and convergent inference for dense random fields. In *International Conference on Machine Learning (ICML)*. 513–521.

Y. Lecun, L. Bottou, Y. Bengio, and P. Haffner. 1998. Gradient-based learning applied to document recognition. *Proc. IEEE* 86, 11 (1998), 2278–2324.

Yangyan Li, Soeren Pirk, Hao Su, Charles R. Qi, and Leonidas J. Guibas. 2016. FPNN: field probing neural networks for 3D data. In *Neural Information Processing Systems (NIPS)*.

Sergey Ioffe and Christian Szegedy. 2015. Batch Normalization: accelerating deep network training by reducing internal covariate shift. In *International Conference on Machine Learning (ICML)*. 448–456.

Jonathan Long, Evan Shelhamer, and Trevor Darrell. 2015. Fully convolutional models for semantic segmentation. In *Computer Vision and Pattern Recognition (CVPR)*.

Jonathan Masci, Davide Boscaini, Michael M. Bronstein, and Pierre Vandergheynst. 2015. Geodesic convolutional neural networks on Riemannian manifolds. In *International Conference on Computer Vision (ICCV)*.

D. Maturana and S. Scherer. 2015. VoxNet: a 3D convolutional neural network for real-time object recognition. In *International Conference on Intelligent Robots and Systems (IROS)*.

Donald Meagher. 1982. Geometric modeling using octree encoding. *Computer Graphics and Image Processing* 19 (1982), 129–147.

Hyeonwoo Noh, Seunghoon Hong, and Bohyung Han. 2015. Learning deconvolution network for semantic segmentation. In *International Conference on Computer Vision (ICCV)*. 1520–1528.

Charles R. Qi, Hao Su, Kaichun Mo, and Leonidas J. Guibas. 2017. PointNet: Deep learning on point sets for 3D classification and segmentation. In *Computer Vision and Pattern Recognition (CVPR)*.

Charles Ruizhongtai Qi, Hao Su, Matthias Nießner, Angela Dai, Mengyuan Yan, and Leonidas J. Guibas. 2016. Volumetric and multi-view CNNs for object classification on 3D data. In *Computer Vision and Pattern Recognition (CVPR)*.

Gernot Riegler, Ali Osman Ulusoy, and Andreas Geiger. 2017. OctNet: Learning deep 3D representations at high resolutions. In *Computer Vision and Pattern Recognition (CVPR)*.

M. Savva, F. Yu, Hao Su, M. Aono, B. Chen, D. Cohen-Or, W. Deng, Hang Su, S. Bai, X. Bai, N. Fish, J. Han, E. Kalogerakis, E. G. Learned-Miller, Y. Li, M. Liao, S. Maji, A. Tatsuma, Y. Wang, N. Zhang, and Z. Zhou 4. 2016. SHREC'16 Track – Large-scale 3D shape retrieval from ShapeNet Core55. In *Eurographics Workshop on 3D Object Retrieval*.

B. Shi, S. Bai, Z. Zhou, and X. Bai. 2015. DeepPano: deep panoramic representation for 3-D shape recognition. *IEEE Signal Processing Letters* 22, 12 (2015), 2339–2343.

Ayan Sinha, Jing Bai, and Karthik Ramani. 2016. Deep learning 3D shape surfaces using geometry images. In *European Conference on Computer Vision (ECCV)*. 223–240.

Nitish Srivastava, Geoffrey E Hinton, Alex Krizhevsky, Ilya Sutskever, and Ruslan Salakhutdinov. 2014. Dropout: a simple way to prevent neural networks from overfitting. *Journal of Machine Learning Research* 15, 1 (2014), 1929–1958.

H. Su, S. Maji, E. Kalogerakis, and E. Learned-Miller. 2015. Multi-view convolutional neural networks for 3D shape recognition. In *International Conference on Computer Vision (ICCV)*.

Jane Wilhelms and Allen Van Gelder. 1992. Octrees for faster isosurface generation. *ACM Trans. Graph.* 11, 3 (1992), 201–227.

Z. Wu, S. Song, A. Khosla, F. Yu, L. Zhang, X. Tang, and J. Xiao. 2015. 3D ShapeNets: A deep representation for volumetric shape modeling. In *Computer Vision and Pattern Recognition (CVPR)*.

Li Yi, Vladimir G. Kim, Duygu Ceylan, I-Chao Shen, Mengyan Yan, Hao Su, Cewu Lu, Qixing Huang, Alla Sheffer, and Leonidas Guibas. 2016. A scalable active framework for region annotation in 3D shape collections. *ACM Trans. Graph. (SIGGRAPH ASIA)* 35, 6 (2016), 210:1–210:12.

Li Yi, Hao Su, Xingwen Guo, and Leonidas Guibas. 2017. SyncSpecCNN: synchronized spectral CNN for 3D shape segmentation. In *Computer Vision and Pattern Recognition (CVPR)*.

Matthew D. Zeiler and Rob Fergus. 2014. Visualizing and understanding convolutional networks. In *European Conference on Computer Vision (ECCV)*.

Kun Zhou, Minmin Gong, Xin Huang, and Baining Guo. 2011. Data-parallel octrees for surface reconstruction. *IEEE. T. Vis. Comput. Gr.* 17, 5 (2011), 669–681.